\documentclass{article}
\usepackage[utf8]{inputenc}
\usepackage{spconf,amsmath,graphicx}
\usepackage{amsfonts}
\usepackage{booktabs}
\usepackage{pifont}
\usepackage{xcolor}
\usepackage[most]{tcolorbox}
\definecolor{mygreen}{cmyk}{0.7, 0, 0.8, 0.2}
\definecolor{thermal}{RGB}{255,165,0} 

\usepackage[normalem]{ulem}

 \usepackage{hyperref}
 \usepackage{wrapfig}
 \usepackage{placeins}

\usepackage{multirow}
\usepackage{comment}
\usepackage{float}
\usepackage{stfloats}  

\newcommand{\cmark}{\ding{51}} 
\newcommand{\xmark}{\ding{55}} 

\definecolor{thermalred}{RGB}{220,50,47}
\definecolor{thermalblue}{RGB}{38,139,210}
\definecolor{thermalgray}{gray}{0.95}
\definecolor{rgbgreen}{RGB}{34,139,34}

\title{See the past: Time-Reversed Scene Reconstruction from Thermal Traces Using Visual Language Models} \vspace{-1em}
\name{Kebin Contreras$^{\star}$ \qquad
Luis Toscano-Palomino$^{\diamondsuit}$ \qquad
Mauro Dalla Mura$^{\dagger\ddagger}$ \qquad
Jorge Bacca$^{\diamondsuit}$$^{\dagger}$ \thanks{This work was partially supported by project ANR-23-IACL-0006 and ECOS Nord project n. C24M01.} }\vspace{-1em}

\address{$^{\star}$ Physics School, Universidad Industrial de Santander, Colombia.\\  $^{\diamondsuit}$Department of Computer Science, Universidad Industrial de Santander, Colombia.\\
$^{\dagger}$ GIPSA-Lab, Université Grenoble Alpes, CNRS, Grenoble INP, Grenoble, France.\\
$^{\ddagger}$ Institut Universitaire de France (IUF), France. \vspace{-1em}}

\begin{document}

\maketitle

\begin{abstract}
Recovering the past from present observations is an intriguing challenge with potential applications in forensics and scene analysis. Thermal imaging, operating in the infrared range, provides access to otherwise invisible information. Since humans are typically warmer ($\approx$ 37 °C / 98.6 °F) than their surroundings, interactions such as sitting, touching, or leaning leave residual heat traces. These fading imprints serve as passive temporal codes, allowing for the inference of recent events that exceed the capabilities of RGB cameras.
We propose a time-reversed reconstruction framework that uses paired RGB and thermal images to recover scene states from a few seconds earlier. The proposed approach couples Visual–Language Models (VLMs) with a constrained diffusion process, where one VLM generates scene descriptions and another guides image reconstruction, ensuring semantic and structural consistency. The method is evaluated on a purpose-built dataset consisting of controlled human–environment interactions. Experimental results demonstrate the feasibility of reconstructing plausible RGB scene states up to 120 seconds prior to observation, establishing a first step toward practical time-reversed imaging based on thermal traces.
\end{abstract}

\section{Introduction} \vspace{-1em}
\label{sec:intro}
Thermal cameras measure long-wave infrared radiation ($\approx8-14 \mu m$), capturing temperature distributions rather than reflected visible light~\cite{vollmer2018infrared}. Unlike RGB sensors, which record instantaneous intensity values in the visible range, thermal imaging provides access to heat transfer processes that often persist after an interaction has ended. These properties have made thermal imaging valuable in applications such as surveillance~\cite{davis2007background} and medical monitoring~\cite{ring2012infrared}.

\begin{figure}[htb]
\centering
\centerline{\includegraphics[width=\linewidth]{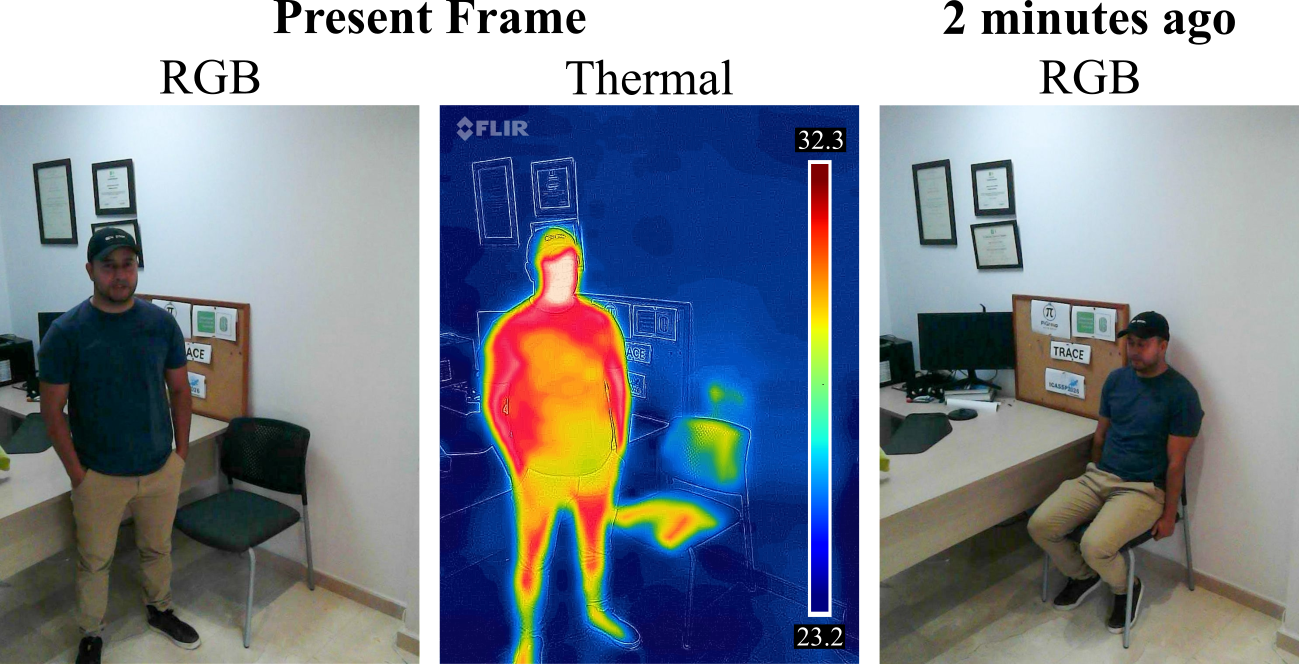}} \vspace{-0.5em}
\caption{\small Illustrative example of residual thermal traces. (\textbf{\textit{Left}}) Present RGB and thermal images, where thermal imprints on the chair indicate a prior human interaction. (\textbf{\textit{Right}}) Ground-truth frame captured approximately two minutes earlier.\vspace{-1em}}
\label{fig:main_figure}
\end{figure}

Humans typically maintain a body temperature of about 37 °C / 98.6 °F, which is higher than that of most surrounding objects~\cite{moran2002core}. Physical contact with chairs, walls, or handheld items leaves behind residual heat patterns that decay over seconds to minutes, depending on the material properties and environmental conditions. These fading imprints function as passive temporal codes, providing not only information about the current scene but also indirect evidence of recent past events. While imperceptible in RGB imagery, such traces are readily observable in the thermal domain. As illustrated in Fig.~\ref{fig:main_figure}, the residual heat marks intuitively indicate a prior interaction, giving observers a hint of “what just happened''.

Most prior work in temporal reasoning has concentrated on predicting the future from observed frames. RGB Video frame extrapolation methods such as Deep Voxel Flow~\cite{liu2017video} synthesize intermediate frames by estimating spatiotemporal voxel flows, while recurrent architectures such as PredNet~\cite{lotter2016deep} and CrevNet~\cite{yu2019crevnet} propagate latent scene representations forward in time to forecast subsequent frames. Although designed for prediction, these models demonstrate that temporal structure can be learned and, if the conditioning were reversed, could in principle be adapted to infer plausible past sequences. However, inference methods based solely on RGB remain highly ambiguous, as many different past trajectories may be equally consistent with the same observed frames. To reduce this uncertainty, residual cues from other modalities have been explored. For example, in~\cite{tang2023happened} an RGB–thermal dataset was introduced in which residual heat imprints enabled recovery of a person’s pose up to three seconds earlier. While promising, this work is limited to pose-level inference, short temporal horizons, and does not address the reconstruction of full-scene appearance. To the best of our knowledge, no previous work has directly tackled the problem of reconstructing complete past scene states from thermal traces.

In this paper, a proof-of-concept for time-reversed scene reconstruction using paired thermal and RGB inputs is presented. The proposed framework combines Visual–Language Models (VLMs) with a constrained diffusion process: one VLM generates semantic descriptions of the scene conditioned on RGB and thermal inputs, while these descriptions guide the diffusion model to synthesize a plausible past frame. By exploiting faded thermal signatures as passive temporal codes and RGB as spatial context, the proposed method reconstructs visually consistent past images from present observations. To support this study, we introduce TraceDataset, a multimodal dataset comprising 80 controlled human–environment interaction scenes, including actions such as sitting on a chair, leaning against a wall, and touching an object. Reconstructions are assessed using both low-level metrics (PSNR, SSIM) and high-level metrics (pose estimation, object detection). The results show that even in controlled scenarios, plausible recovery of past frames is achievable, marking the first step toward scene-level time-reversed imaging.
\section{Methodology}

\subsection{Problem Formulation}
The problem of time-reversed scene reconstruction is formulated as the task of inferring a plausible past image $\mathbf{x}^{t-\Delta}_{RGB}$ from current static multimodal observations. Specifically, access is assumed to an RGB frame $\mathbf{x}_{RGB}^t \in \mathbb{R}^{h \times w \times 3}$ and a co-registered thermal measurement $\mathbf{x}_{Thermal}^t \in \mathbb{R}^{h \times w}$ capturing residual heat traces left by human–environment interactions. The objective is to exploit these complementary modalities to recover a physically consistent estimate of the scene $\hat{\mathbf{x}}^{t-\Delta}_{RGB}$ at a short time horizon $\Delta > 0$.

\subsection{Proposed Method}

A framework is proposed for time-reversed scene reconstruction that leverages residual thermal traces, combined with RGB imagery for spatial context. The method consists of three main components: (i) multimodal input encoding, (ii) VLM guidance, and (iii) a constrained diffusion backbone for past-frame generation.

\subsubsection{Multimodal Input Encoding}

Given the current scene, a co-registered RGB image $\mathbf{x}_{RGB}^t$ and a thermal image $\mathbf{x}_{Thermal}^t$ are acquired at time $t=0$. The RGB channel provides spatial detail, while the thermal modality encodes fading heat signatures caused by recent human--environment interactions. These two inputs are normalized and aligned before being passed to the VLM pipeline.

\subsubsection{VLM-Based Guidance}

The first step of the proposed method consists in producing a structured textual description of the current scene, conditioned on the multimodal inputs $(\mathbf{x}_{RGB}^t, \mathbf{x}_{Thermal}^t)$ and the textual prompt $p_{desc}$. The resulting output $p_{out}$ encodes high-level semantics (e.g., \texttt{``a person was recently sitting on the chair; residual heat is visible''}), which serve as contextual cues to guide subsequent reconstruction. Importantly, we do not train a new model on RGB--thermal--text triplets. Instead, an existing pretrained VLM (GPT-5) is employed in inference mode, where the parameters $\theta^*$ remain frozen. This reduces the operation to a forward pass without gradient updates:

\begin{equation}
p_{out} = \arg\max_{y} \; p\big(y \mid \mathbf{x}_{RGB}^t, \mathbf{x}_{Thermal}^t, p_{desc}; \theta^*\big).
\end{equation}
Here, $p_{out}$ denotes the most likely textual description generated by the off-the-shelf VLM, given the RGB frame $\mathbf{x}_{RGB}^t$, the thermal frame $\mathbf{x}_{Thermal}^t$, and the conditioning prompt $p_{desc}$, under the frozen parameters $\theta^*$.

\subsubsection{Constrained Diffusion Framework}

Then, the desired past image is generated using an existing pretrained diffusion model (Gemini 2.5 Flash Image, also known as NanoBanana), which was originally trained on large-scale general-purpose image datasets. In this method, the diffusion model is neither retrained nor finetuned; instead, its denoising trajectory is constrained by conditioning jointly on the multimodal inputs, the structured scene description $p_{out}$, and the generative prompt $p_{gen}$. This guidance ensures that the reconstruction remains semantically and physically consistent with both the observed modalities and the textual priors.

Formally, let $\textbf{z}_T \sim \mathcal{N}(0,I)$ denote the initial Gaussian noise in the latent space. The pretrained diffusion-based generator produces a plausible past frame by iteratively refining this latent representation undermultimodal conditioning:
\begin{equation}
\textbf{z}_{t-1} = g(\textbf{z}_t, \mathbf{x}_{RGB}^t, \mathbf{x}_{Thermal}^t,
p_{out}, p_{gen}), \quad t = T,\dots,1.
\end{equation}
where $g$ denotes the implicit denoising operator of the pretrained diffusion-based image generator. In our setting, reconstruction is achieved in a plug-and-play manner using an off-the-shelf commercial diffusion model, by conditioning on the current RGB observation  with fixed textual prompt templates, which provide semantic and physical constraints derived from residual heat traces.
The final reconstructed past frame is directly obtained from the commercial diffusion-based generator output, yielding an RGB estimate conditioned on the observed thermal traces and the semantic descriptor as, 
\begin{equation}
\hat{\mathbf{x}}^{t-\Delta t}_{RGB} =
G(\mathbf{x}_{RGB}^t, \mathbf{x}_{Thermal}^t, p_{out}, p_{gen}),
\label{equation:_final}
\end{equation}
where Eq.~\ref{equation:_final} represents the constrained generation of a plausible past scene $\hat{\mathbf{x}}^{t-\Delta t}_{RGB}$, obtained purely through inference with frozen parameters $\theta^*$, where the prompts $p_{out}$ and $p_{gen}$ act as conditioning signals to guide the pretrained model toward reconstructions aligned with physically consistent semantics. \vspace{-2em}
\begin{figure*}
    \centering
    \includegraphics[width=1\linewidth]{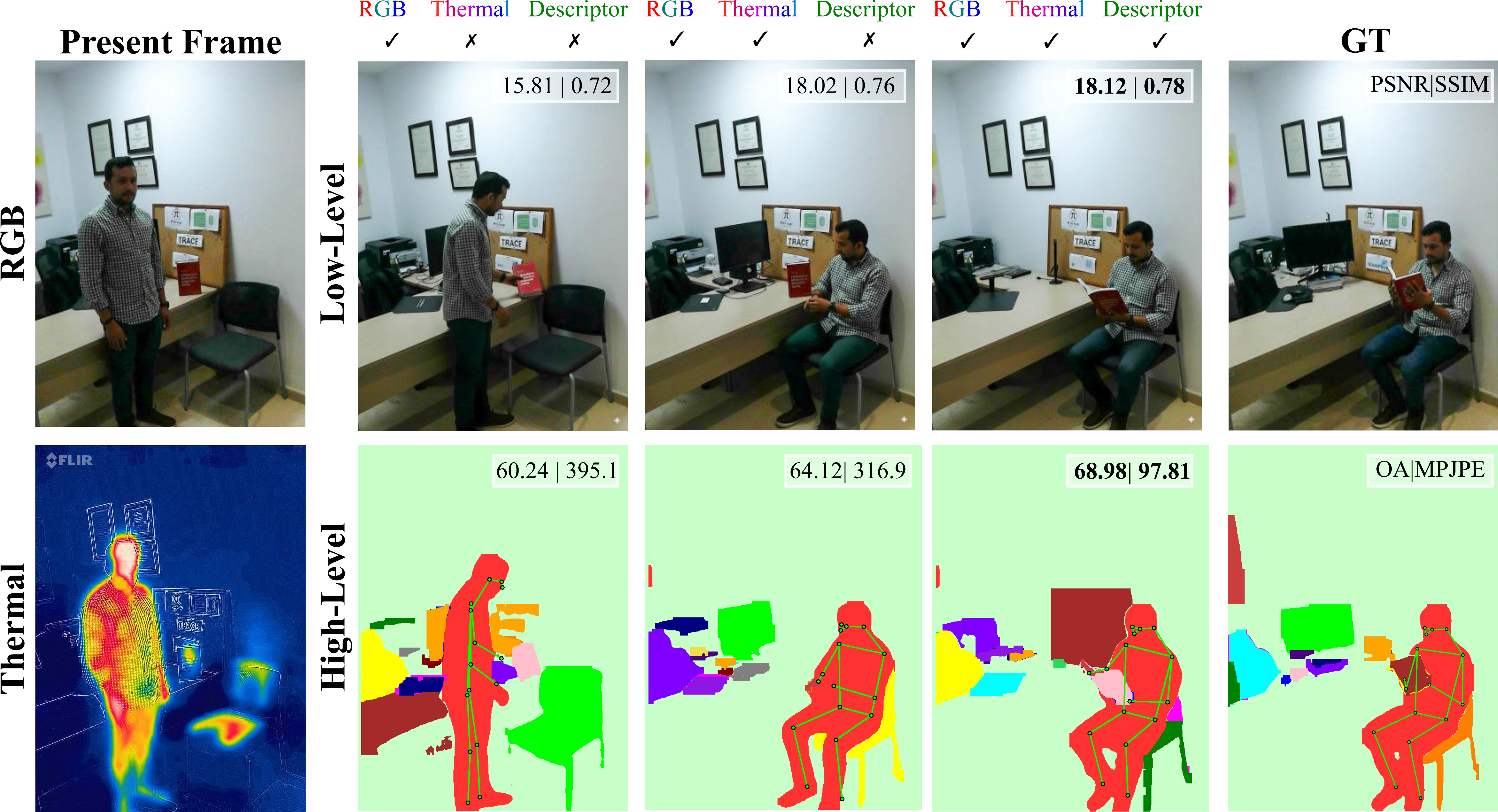}\vspace{-1.5em}
    \caption{\small Ablation study comparing RGB only, RGB+Thermal, and RGB+Thermal+Descriptor against ground truth, evaluated using low-level metrics (PSNR, SSIM) and high-level metrics (semantic segmentation with OA, pose estimation with MPJPE) to see the past, 30 seconds ago.} \vspace{-0.5em}
    \label{fig:cross_domain_visual}
\end{figure*}
\section{Simulations and Results}
\vspace{-0.5em}

We introduce the TRACE dataset to evaluate our method, comprising 80 multimodal scenes centered on common physical interactions such as \textit{sitting on a chair}, \textit{touching an object}, and \textit{leaning against a wall}. In each case, a person maintained physical contact with the scene for 30 seconds, after which RGB and thermal images were acquired at multiple time delays ($5s, 15s, 30s, 120s$). Image acquisition was performed using a FLIR ONE camera, which provides aligned RGB and Long-Wave InfraRed thermal images. The RGB sensor has a resolution of $1440 \times 1080$, while the thermal sensor has a resolution of $80 \times 60$, operating at a frame rate of 8.7 Hz with a horizontal field of view of $55^\circ$. The acquisition at $0s$ was taken as the ground-truth reference. Reconstructions of past frames were evaluated against the ground truth using both low-level and high-level metrics. For low-level metrics, the Peak Signal-to-Noise Ratio (PSNR) and the Structural Similarity Index Measure (SSIM) were implemented according to~\cite{monroy2025generalized}. For high-level metrics, pose estimation with the Mean Per Joint Position Error (MPJPE)~\cite{tang2023happened} and semantic segmentation with the Overall Accuracy (OA)~\cite{hinojosa2018coded} were employed.

The proposed method is composed of two vision–language models (VLMs): GPT-5, used to generate structured scene descriptions, and Gemini 2.5 Flash Image (NanoBana), employed as the conditional diffusion model for image synthesis. A comparative analysis of different VLMs is presented in Section 3.2.

The method relies on two fixed prompts: one for structured scene description and another for prompt generation for the diffusion model, as detailed below. These prompts were derived from a systematic prompt analysis described in the Supplementary Material (\href{https://drive.google.com/drive/folders/13cjV_KshOzF92Tg0YO7VDrTZkE4S4vdp?usp=sharing}{SM } ).

\begin{tcolorbox}[
  enhanced,
  colback=thermalgray,
  colframe=rgbgreen!70!thermalblue,
  coltitle=black,
  fonttitle=\bfseries,
  title=Prompt for Scene Description ($p_{desc}$),
  boxrule=0.8pt,
  arc=3mm,
  left=2mm, right=2mm, top=1mm, bottom=1mm
]
\begin{itemize}
    \item \texttt{Identify the thermographic traces on the thermal image.}
    \item \texttt{In the RGB image, describe the scene. } 
    \item \texttt{Output one concise past-tense sentence describing what happened some seconds or minutes ago, including all these details. } 
\end{itemize}
\end{tcolorbox}

\begin{tcolorbox}[
  enhanced,
  colback=thermalgray,
  colframe=thermalred!80!thermalblue,
  coltitle=black,
  fonttitle=\bfseries,
  title=Prompt for Generative Editing ($p_{gen}$),
  boxrule=0.8pt,
  arc=3mm,
  left=2mm, right=2mm, top=1mm, bottom=1mm
]
\texttt{Edit the RGB image using <description $p_{out}$> or <the thermal image> to depict the scene a few seconds earlier, preserving environment, lighting, angle, zoom, and colors.}
\end{tcolorbox}

Four main experiments were conducted:  (i) an ablation study evaluating the contribution of the main components of the method,  (ii) a state-of-the-art comparison between image generators, (iii) an assessment of how far into the past the proposed approach can reliably reconstruct in the controlled scenarios and (iv) a prompt analysis found in \href{https://drive.google.com/drive/folders/13cjV_KshOzF92Tg0YO7VDrTZkE4S4vdp?usp=sharing}{SM } .  \vspace{-1em}

\subsection{Ablation Study} \vspace{-0.5em}

To evaluate the impact of each input modality and the contribution of the scene descriptor within the overall framework, an ablation was conducted. Table~\ref{tab:ablacion} summarizes both high-level and low-level metrics under three configurations: (i) RGB input only, (ii) RGB combined with thermal input without the descriptor, and (iii) the full model incorporating RGB, thermal, and descriptor guidance. The results indicate a significant improvement when thermal information is incorporated. However, VLMs alone struggle to correctly interpret residual traces to infer past events if a structured semantic description is not provided. When the descriptor is included, performance increases by 10\% in AO and 1.3 dB in PSNR. This effect is illustrated in Fig.~\ref{fig:cross_domain_visual}, where the qualitative results are consistent with the numerical trends. Using RGB-only inputs fails to reveal the previous state. Adding thermal data introduces complementary cues, allowing recovery of the person’s prior action of sitting down a few seconds earlier. The most significant improvement, however, occurs when incorporating the scene descriptor: this semantic prior enables the model to infer higher-level interactions, such as detecting that the book was in hand. These results confirm that the VLM-based structured description acts as a strong prior, anchoring the diffusion process to physically consistent semantics. 

\begin{table}[!b]
  \centering
    \caption{\small Ablation study with different input modalities for the case of 30 seconds ago. \vspace{1em}} 
  \label{tab:ablacion}
  \renewcommand{\arraystretch}{1.3}
  \resizebox{\columnwidth}{!}{
  \begin{tabular}{ccc|cc|cc}
    \toprule
    \textcolor{red}{R}\textcolor{mygreen}{G}\textcolor{blue}{B} & 
    
    \textcolor{red}{T}%
    \textcolor{red!70!blue}{h}%
    \textcolor{red!55!blue}{e}%
    \textcolor{red!40!blue}{r}%
    \textcolor{red!25!blue}{m}%
    \textcolor{red!10!blue}{a}%
    \textcolor{blue}{l} 
    
    & \textcolor{rgbgreen!70!thermalblue}{\textbf{Descriptor}}

    & \multicolumn{2}{c|}{\textbf{High-level}} 
    & \multicolumn{2}{c}{\textbf{Low-level}} \\ 
    \cmidrule(lr){4-5} \cmidrule(lr){6-7}
     & & & \textbf{AO ($\uparrow$)} & \textbf{MPJPE ($\downarrow$)} & \textbf{PSNR ($\uparrow$)} & \textbf{SSIM ($\uparrow$)} \\ 
    \midrule

    \cmark & \xmark & \xmark & 62.24 $\pm$ 0.75 & 336.92 $\pm$ 98.18 & 16.81 $\pm$ 0.54 & 0.76 $\pm$ 0.65 \\

    \cmark & \cmark & \xmark & 67.55 $\pm$ 0.13 & 270.87 $\pm$ 13.02 & 17.22 $\pm$ 1.08 & 0.77 $\pm$ 0.14 \\

    \cmark & \cmark & \cmark & \textbf{76.87 $\pm$ 0.06} & \textbf{48.44 $\pm$ 14.18} & \textbf{18.57 $\pm$ 1.21} & \textbf{0.79 $\pm$ 0.02} \\
    \bottomrule
  \end{tabular}
  }
\end{table}

\subsection{Image Generator Comparison}

\begin{figure}[!t]
\centering
\centerline{ \includegraphics[width=\columnwidth]{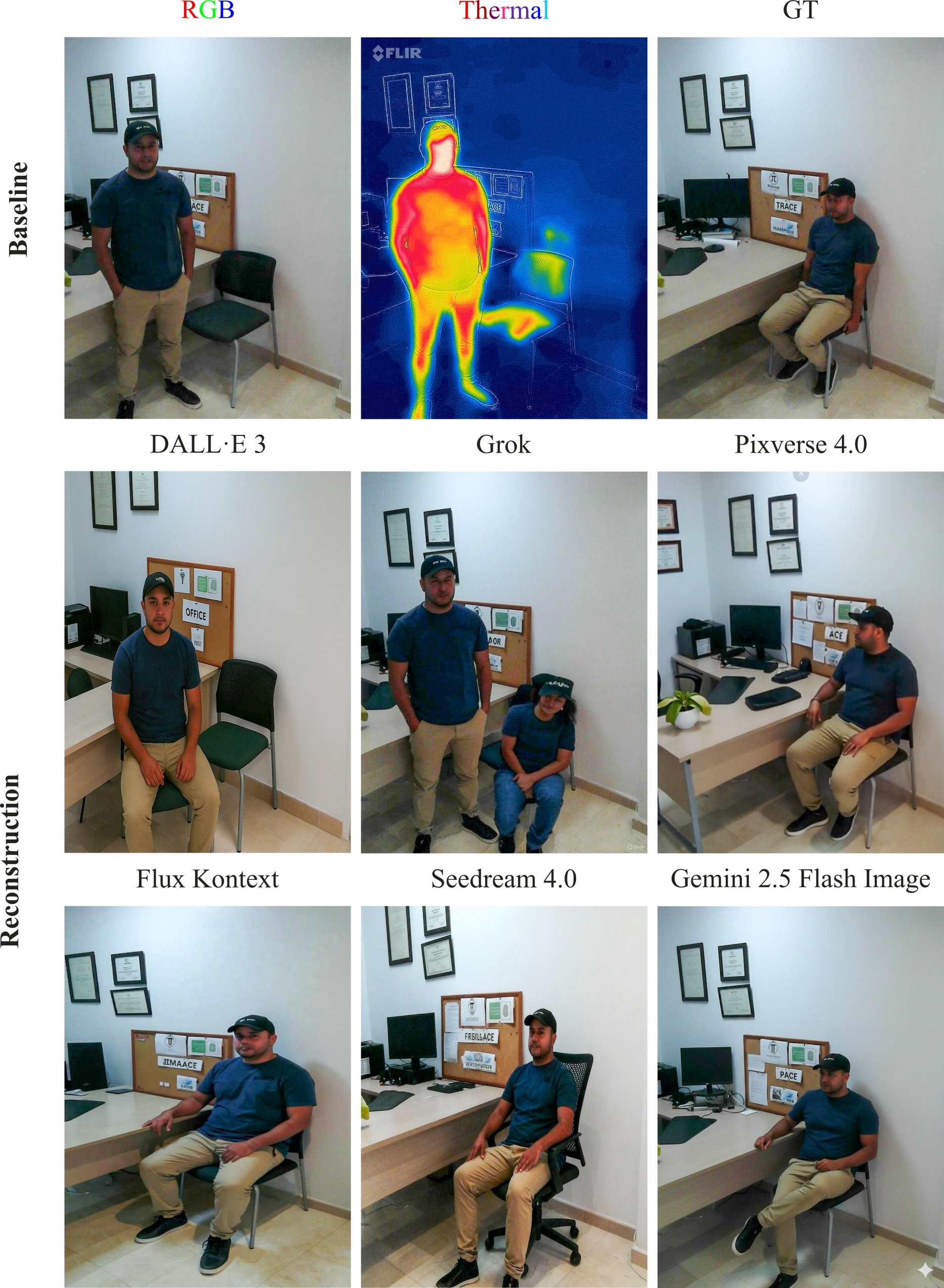}}
    \caption{  \small 
Comparison between the image generators to estimate the past. 
The first row shows the inputs (RGB) and thermal, and the ground-truth (GT) scene. 
The subsequent rows illustrate past-scene reconstruction produced by different generative models.} 
\label{fig:model_analysis} 
\end{figure}

To investigate how different generative models perform in past-scene reconstruction, we employed the proposed scheme and only changed the generative model. For that we compared with DALLE 3 \cite{openai2023dalle3}, Grok~\cite{xai2025grok}, Pixverse~\cite{pixverse2025}, Flux Kontext\cite{batifol2025flux}, Seeddream 4.0~\cite{seedream2025seedream}, and Gemini 2.5 Flash Image~\cite{google2025gemini2_5}. All of them are VLM, where the generating process is constrained by the prompt used in our methodology. Table \ref{tab:comparativa_metodos} presents the quantitative results, demonstrating that Gemini 2.5 Flash achieves the highest performance and most consistent results in almost all evaluated metrics. To further illustrate these findings, a qualitative comparison is provided in Figure~\ref{fig:model_analysis}: the top row shows the reference RGB and thermal input image, while the following rows display the reconstruction produced by different generative models. This setup allows us to systematically assess how models vary in their ability to integrate multimodal cues, preserve environmental fidelity, and render temporally consistent human actions, thereby shedding light on their robustness in tasks that require reasoning about the immediate past. 

\begin{table}[!t]
  \centering
  \caption{\small Comparison of our proposed Gemini 2.5 Flash Image framework against state-of-the-art generative models for time-reversed reconstruction (30s delay). MPJPE values are reported scaled by $\times 10^{5}$.}  \vspace{1em}
  \label{tab:comparativa_metodos}
  \renewcommand{\arraystretch}{1.3}
  \resizebox{\columnwidth}{!}{
  \begin{tabular}{l|cc|cc}
    \toprule
    \multirow{2}{*}{\textbf{Method}} & \multicolumn{2}{c|}{\textbf{High-level (Semantic)}} & \multicolumn{2}{c}{\textbf{Low-level (Structural)}} \\ 
    \cmidrule(lr){2-3} \cmidrule(lr){4-5}
     & \textbf{AO ($\uparrow$)} & \textbf{MPJPE ($\downarrow$)} & \textbf{PSNR ($\uparrow$)} & \textbf{SSIM ($\uparrow$)} \\ 
    \midrule
    DALL-E 3 \cite{openai2023dalle3} & 78.5 & 70.2 & 13.74 & 0.53 \\
    Grok \cite{xai2025grok} & 54.4 & 46.6 & 13.38 & 0.52 \\
    Pixverse \cite{pixverse2025} & 80.00 & 67.6 & 14.20 & 0.54 \\
    Flux Kontext \cite{batifol2025flux} & 54.6 & 76.7 & 13.78 & 0.51 \\
    Seeddream 4.0 \cite{seedream2025seedream} & 83.00 & \textbf{11.8} & 15.70 & 0.57 \\
    \midrule
    \textbf{Gemini 2.5 Flash (Ours)} \cite{google2025gemini2_5} & \textbf{85.10} & 74.1 & \textbf{16.00} & \textbf{0.58} \\
    \bottomrule
  \end{tabular}
  }
\end{table}

Importantly, the proposed methodology is inherently model-agnostic. It is not tied to any specific generative architecture, but instead defines a modular framework. Consequently, future advances in generative modeling can be incorporated directly without structural modifications, thereby enabling the framework to benefit naturally from improvements in semantic consistency and physical plausibility.

\begin{figure}[h]
\centering
\centerline{\includegraphics[width=\linewidth]{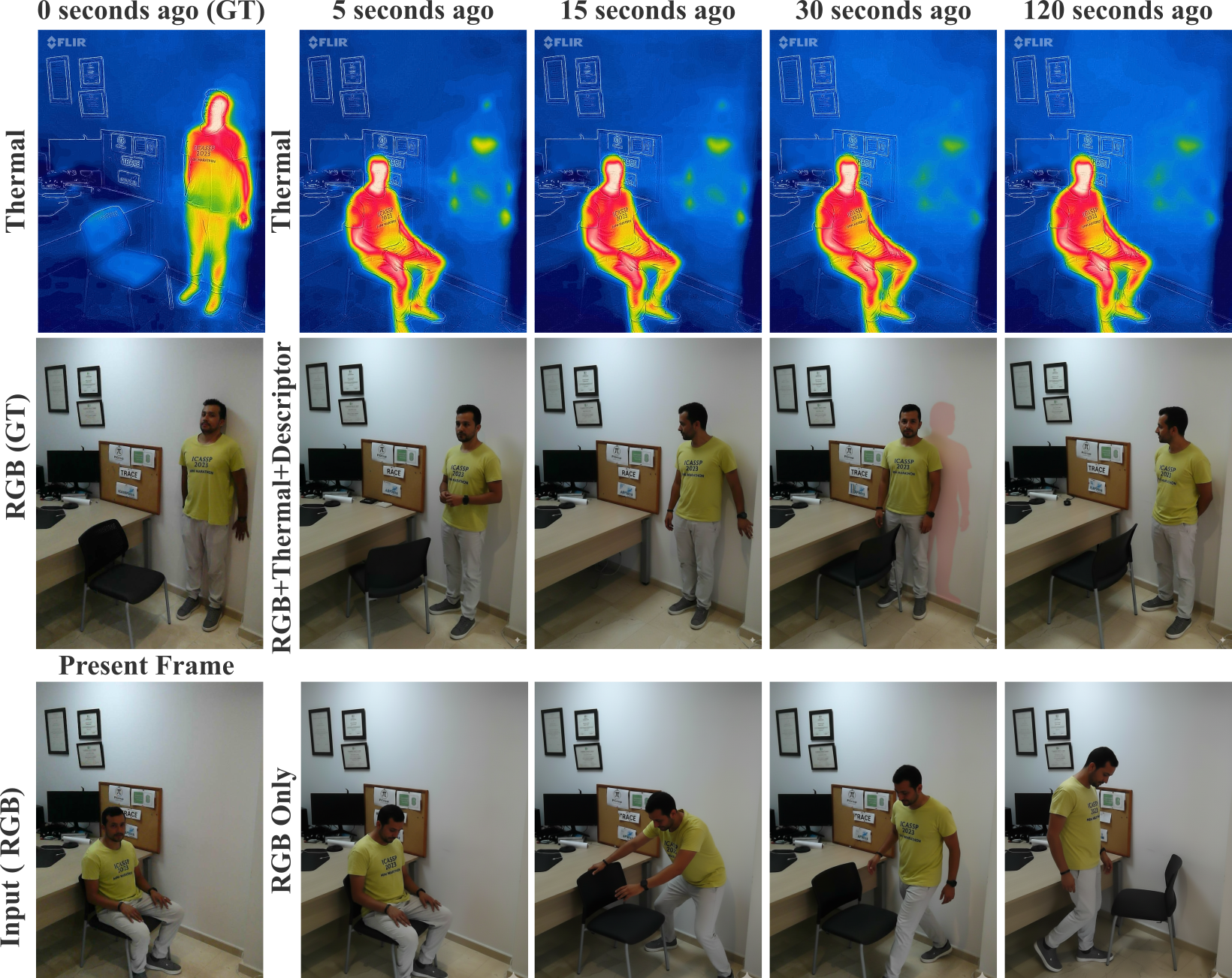}}
\caption{\small Temporal reconstruction at increasing delays (0–120s). Top: thermal frames showing gradual heat decay. Bottom: RGB reconstructions guided by thermal traces and textual priors, recovering plausible past interactions up to two minutes after contact.}
\label{fig:temporal}
\end{figure}

\subsection{Temporal Reconstruction Range} 

The second experiment assesses how far into the past the proposed method can reliably reconstruct a scene. Reconstructions were generated at increasing time delays ($5s, 15s, 30s, $ and $ 120s$) and compared against the ground truth captured at $0s$. Performance was evaluated using both low-level and high-level metrics. As shown in Fig.~\ref{fig:temporal}, the top row depicts thermal frames, where the heat imprints progressively fade over time. The middle row presents the corresponding RGB reconstructions produced by the proposed framework, while the bottom row shows results from the RGB-only baseline. Notably, the RGB-only approach tends to hallucinate plausible but unconstrained frames, since it lacks the guidance provided by thermal information.

\begin{figure}[h]
  \centering
\includegraphics[width=\columnwidth]{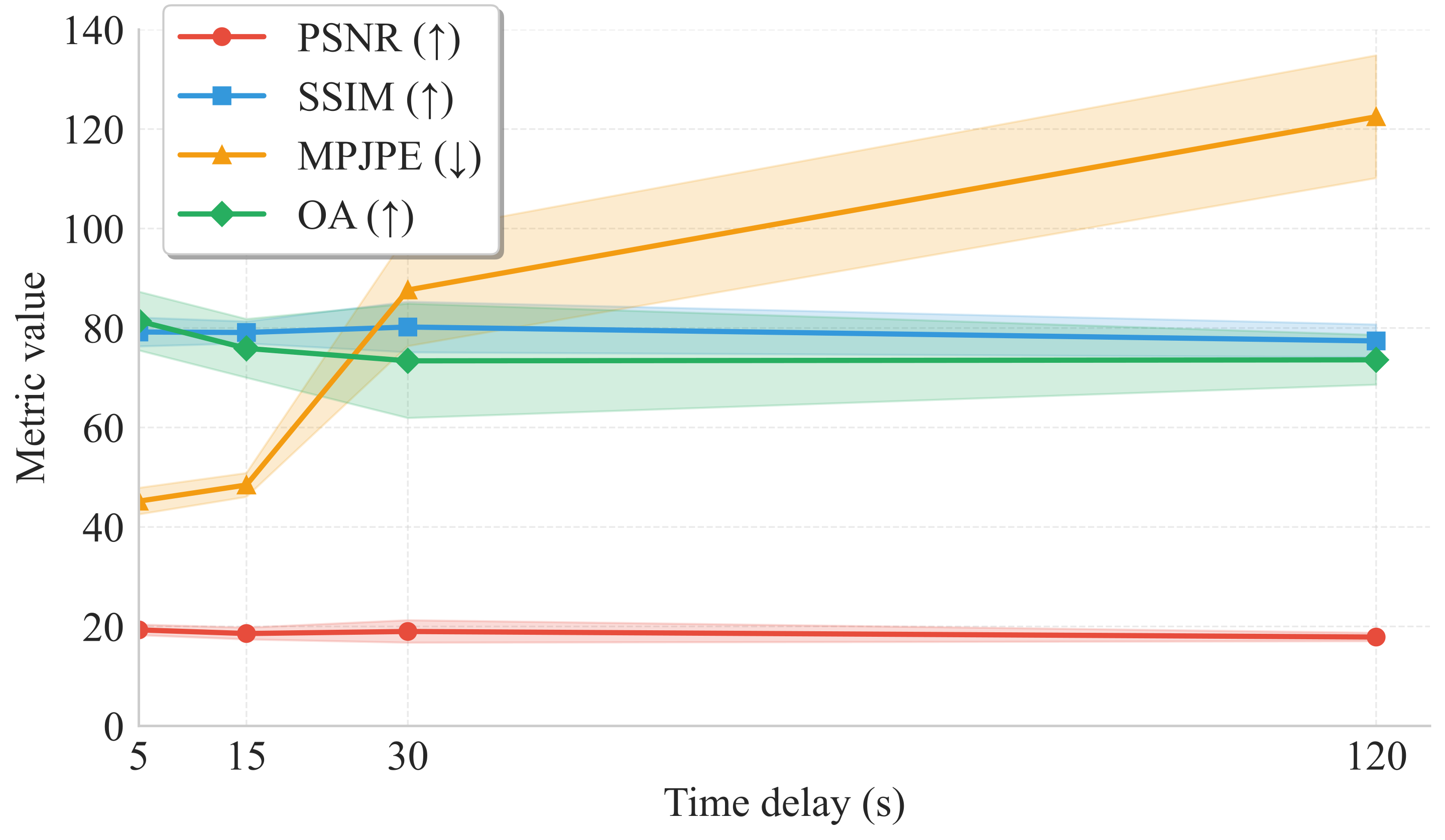}\vspace{-2em}
  \caption{\small Temporal reconstruction performance across increasing time delays. 
  Both low-level metrics (PSNR, SSIM$\times 100$) and high-level metrics (MPJPE, OA) are plotted.}
  \label{fig:temporal_range}
\end{figure}

Figure \ref{fig:temporal_range} summarizes the quantitative results, indicating that the reconstructions remain stable up to $15 s$, with only slight degradation in both fidelity and semantic accuracy. Beyond $30s$, however, the residual thermal traces weaken significantly, leading to noticeable declines in pose estimation and overall accuracy. At $120s$, reconstructions still preserve the global structure of the scene but fail to capture fine details or precise interactions. These findings suggest that the effective temporal horizon of the proposed approach is between one and two minutes, after which the information available in residual traces becomes insufficient for reliable recovery.

\section{Conclusions and Future Work}  \vspace{-0.5em}
This work introduces a fundamentally new perspective on visual scene understanding: recovering recent past states from present observations. By combining RGB and thermal imaging with VLM-guided diffusion models, we present the first framework that explicitly interprets fading thermal imprints as passive temporal codes for reconstructing prior RGB scene configurations. To the best of our knowledge, this is the first study to operationalize residual heat signatures as a computational signal for time-reversed image synthesis. Controlled experiments validate the feasibility of this approach, showing that textual descriptions provide strong guidance and yield consistent gains across both low-level (PSNR, SSIM) and high-level (MPJPE, OA) metrics. The ablation study highlights the importance of VLM-based semantic priors, while temporal analysis demonstrates that past interactions can be plausibly recovered up to two minutes after contact. Future work will extend the framework to real-world environments with multiple subjects and variable illumination, and explore additional sensing modalities and efficiency optimizations for practical deployment in applications such as forensics, human–computer interaction, and security monitoring.

\bibliographystyle{IEEEbib}
\bibliography{main}

\clearpage
\onecolumn
\clearpage
\appendix

\begin{center}
    \section*{See the past: Time-Reversed Scene Reconstruction from Thermal Traces Using Visual Language Models\\(Supplemental Materials)}
    \vspace{-1em}

\end{center}

\vspace{2em}

\begin{wrapfigure}{R}{0.70\textwidth} \vspace{-3em}
    \includegraphics[width=0.65\textwidth]{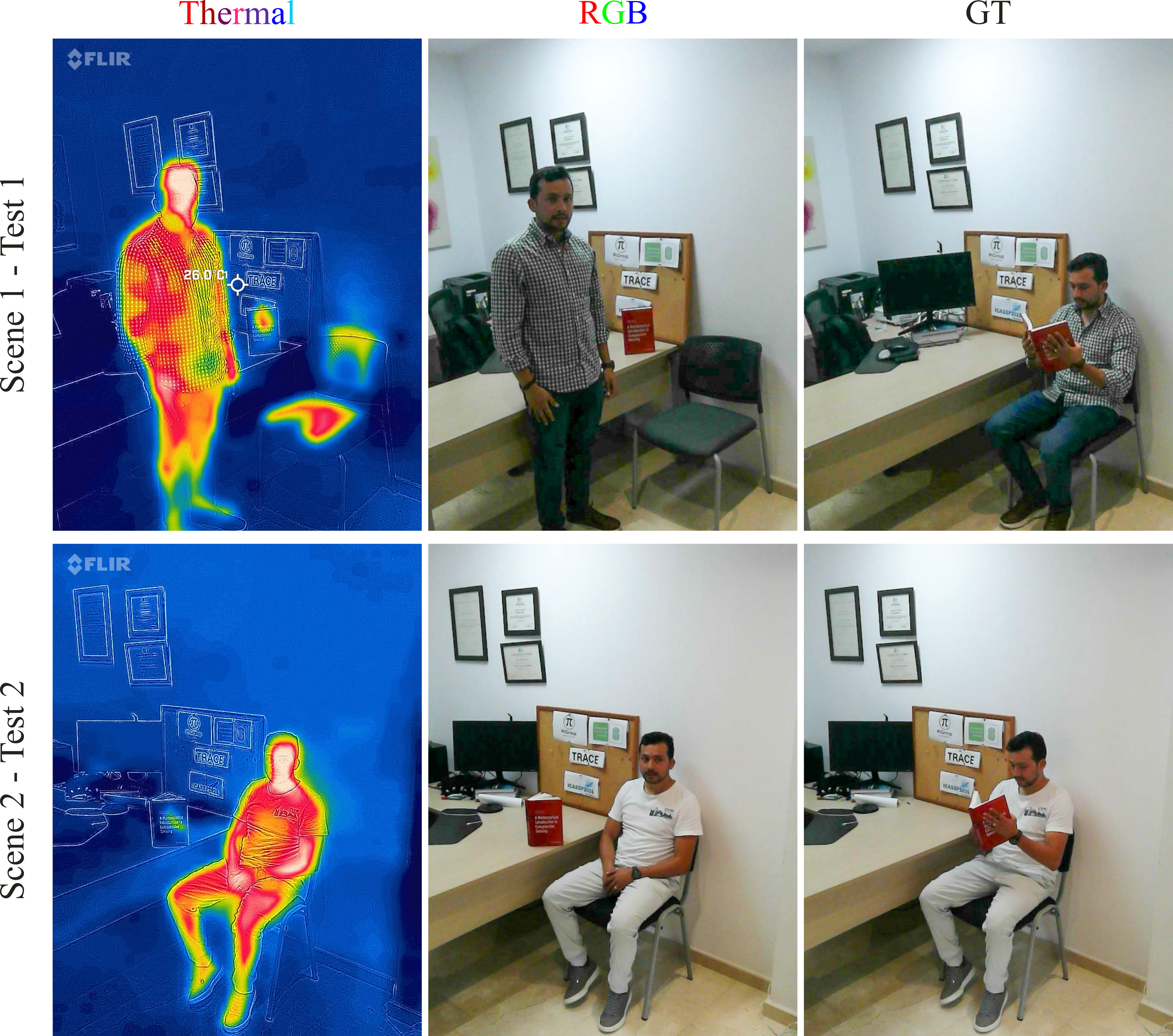} \vspace{-1em}
    \caption{ \small Input and ground-truth data for the supplementary material experiments.    Each test scene includes the thermal image (left) capturing residual heat traces, the corresponding RGB image (middle), and the ground-truth action (right).    Results are illustrated for Scene~1 (Test~1) and Scene~2 (Test~2)} \vspace{-4em}
    \label{fig:sm_inputs} 
\end{wrapfigure}

\vspace{2em} This supplementary material is organized into two sections: (i) Test~1, which analyzes the effect of different prompt descriptors on the accuracy and specificity of past-action reconstruction and (ii) Test~2, which evaluates the role of a prompt editor in refining and structuring the generated descriptions. Figure~\ref{fig:sm_inputs} illustrates the input data used across the experiments. In each test, the model receives paired thermal and RGB images as input, where the thermal modality provides residual heat traces encoding the subject’s recent interactions, and the RGB modality supplies contextual scene information. The ground-truth (GT) frame corresponds to the subject’s actual position and action 20s earlier, serving as the reference for evaluation. 

\par
\bigskip

\begin{center}
\mbox{}
\end{center}
\vspace{0.5em}
\section*{Test 1: Prompts descriptor Analysis}
\begin{figure*}[!ht]
    \centering
    \includegraphics[width=\linewidth]{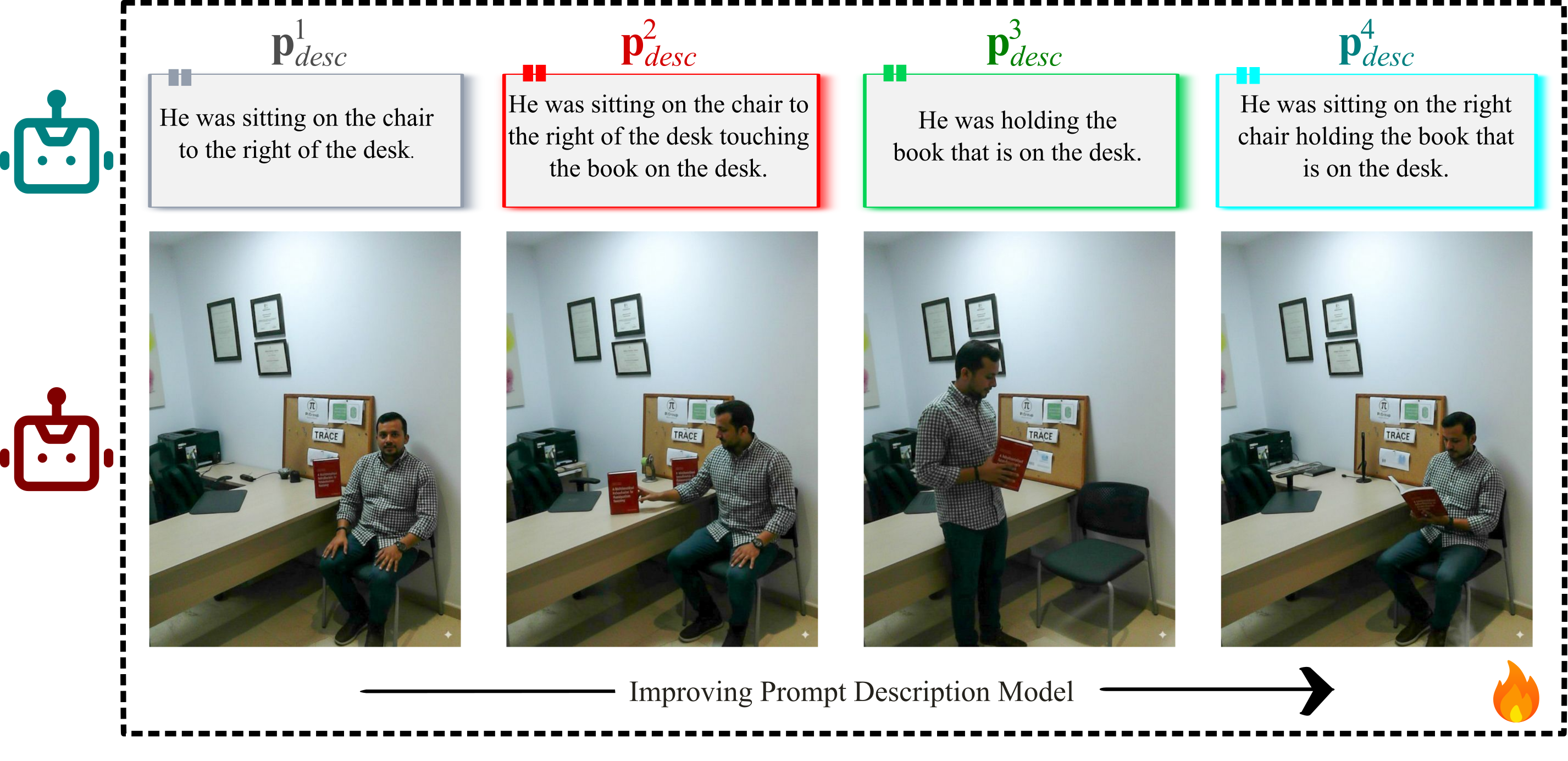} \vspace{-2em}
    \caption{Examples of past-action descriptions generated under different prompt descriptors $p_{\text{desc}}^{1}$--$p_{\text{desc}}^{4}$. 
    The prompts progressively add reasoning constraints, from a simple direct description to structured analyses that incorporate thermal traces, RGB context, and completeness checks. 
    More structured prompts yield descriptions that are more specific and consistent with the evidence.} \vspace{-0.5em}
    \label{fig:prompt-desc}
\end{figure*}

To study how instruction design affects ``what-just-happened'' descriptions from paired RGB--thermal inputs, we evaluate four descriptor prompts $p_{\text{desc}}^{1}$--$p_{\text{desc}}^{4}$ (Fig.~\ref{fig:prompt-desc}). Each prompt must output a single past-tense sentence stating the person’s position and action 5~s earlier, while progressively adding explicit reasoning constraints: $p_{\text{desc}}^{1}$ requests only the final sentence; $p_{\text{desc}}^{2}$ requires enumerating primary/secondary heat traces in the thermal image; $p_{\text{desc}}^{3}$ cross-checks those traces against the RGB image with object attributes and spatial relations; and $p_{\text{desc}}^{4}$ adds a completeness check before synthesizing the final hypothesis. This progression isolates the effect of prompt structure, encouraging grounding on thermal residues and reducing ambiguity or hallucinations. Only the final sentence is returned; intermediate steps are used implicitly to guide inference. Empirically, more structured prompts yield descriptions that are more specific and consistent with the heat-trace evidence, especially in scenes involving multiple interactions (e.g., sitting while holding a book). The exact prompts used in the experiment are shown below, where the differences between the prompts and the previous one are highlighted in bold.

\begin{itemize}
    \item $p_{desc}^1$ = \textit{Follow strictly the step-by-step methodology below and do not skip or shorten any step before giving the final one-sentence answer.An RGB image and a thermal image of the same scene are attached. \textbf{Final output:} Provide only one short, direct sentence in past tense that precisely describes the person’s position and action 20 seconds ago according to the heat traces. The answer must be concise and direct. Output only that sentence, nothing else.}
    \item $p_{desc}^2$ = \textit{Follow strictly the step-by-step methodology below and do not skip or shorten any step before giving the final one-sentence answer.
An RGB image and a thermal image of the same scene are attached.
\textbf{1. Analyze the thermal image and list all objects or furniture that show any heat traces, starting with the one with the highest heat concentration (primary). Do not omit any object that shows heat, even if the trace is faint, small, or low intensity. Explicitly state when an object shows no heat.
2. Identify and describe all secondary heat traces (for example, on books, chairs, walls, desks, or other surfaces). Mention each one individually and describe its approximate intensity (strong, moderate, faint).}
\textbf{Final output:} Provide only one short, direct sentence in past tense that precisely describes the person’s position and action 20 seconds ago, according to the heat traces. The answer must be concise and direct. Output only that sentence, nothing else.}
\item $p_{desc}^3 = $ \textit{Follow strictly the step-by-step methodology below and do not skip or shorten any step before giving the final one-sentence answer.
An RGB image and a thermal image of the same scene are attached.
1. Analyze the thermal image and list all objects or furniture that show any heat traces, starting with the one with the highest heat concentration (primary). Do not omit any object that shows heat, even if the trace is faint, small, or low intensity. Explicitly state when an object shows no heat.
2. Identify and describe all secondary heat traces (for example, on books, chairs, walls, desks, or other surfaces). Mention each one individually and describe its approximate intensity (strong, moderate, faint).
\textbf{3. Cross-check with the RGB image and locate every object with heat traces. For each object, provide:
Object type, Object color, Position (left, center, right), Interaction with the person (touching, sitting, holding, near, none), Direction relative to the scene (front, back, left, right, corner)}
Final output: Provide only one short, direct sentence in past tense that precisely describes the person’s position and action 20 seconds ago according to the heat traces. The answer must be concise and direct. Output only that sentence, nothing else.}
\item  $p_{desc}^4 = $ \textit{Follow strictly the step-by-step methodology below and do not skip or shorten any step before giving the final one-sentence answer.
An RGB image and a thermal image of the same scene are attached.
1. Analyze the thermal image and list all objects or furniture that show any heat traces, starting with the one with the highest heat concentration (primary). Do not omit any object that shows heat, even if the trace is faint, small, or low intensity. Explicitly state when an object shows no heat.
2. Identify and describe all secondary heat traces (for example, on books, chairs, walls, desks, or other surfaces). Mention each one individually and describe its approximate intensity (strong, moderate, faint).
3. Cross-check with the RGB image and locate every object with heat traces. For each object, provide:
Object type, Object color, Position (left, center, right), Interaction with the person (touching, sitting, holding, near, none), Direction relative to the scene (front, back, left, right, corner)
\textbf{4. Before making the final inference, double-check that no object with visible heat traces has been left out of your analysis.
5. Based on both the primary and secondary heat traces, infer the most likely past action of the person 5 seconds ago. If multiple objects have heat traces, combine them into one concise sentence that mentions all relevant actions.}
Final output: Provide only one short, direct sentence in past tense that precisely describes the person’s position and action 20 seconds ago according to the heat traces. The answer must be concise and direct. Output only that sentence, nothing else.}
\end{itemize}

\section*{Test 2: Prompts Editor Analysis}

\begin{figure*}[h]
    \centering
    \includegraphics[width=\linewidth]{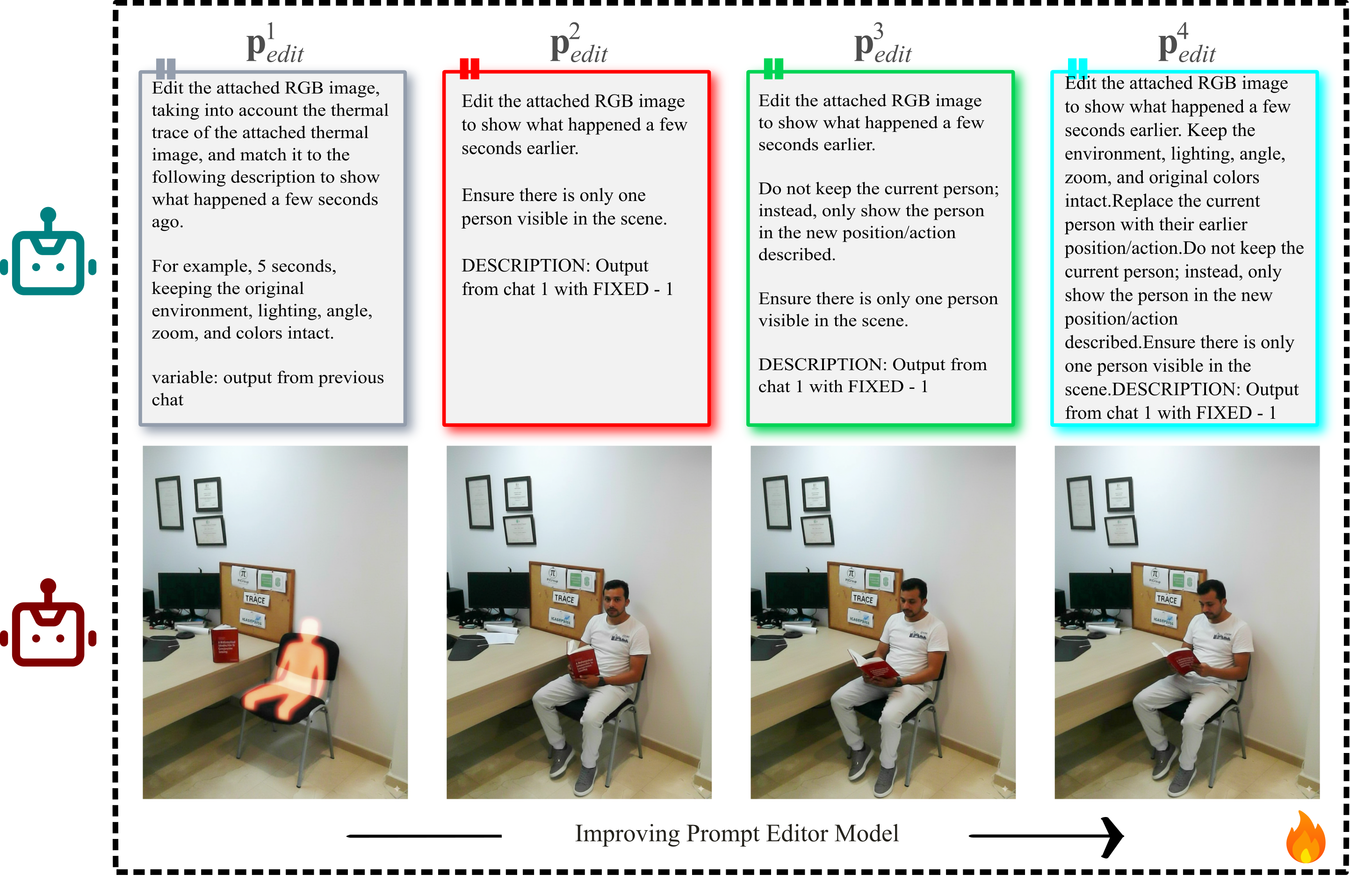} \vspace{-2em}
    \caption{Examples of past-action image predictions generated under different prompt descriptors $p^{1}_{edit}$--$p^{4}_{edit}$. 
    Each prompt introduces progressively stricter editing constraints, from preserving the original environment with minimal changes to enforcing temporal consistency and position replacement. 
    In all cases, the underlying action description used in the prompts was the same: \textit{``the person was sitting and holding the book''}. 
    The structured prompts guide the editing model toward outputs that are more coherent with the thermal trace, scene context, and past actions, demonstrating the impact of prompt design on editing quality and consistency.}
    
    \label{fig:prompt-edition}
\end{figure*}

To study how instruction design affects past-action \textit{predictions} from paired RGB--thermal inputs, we evaluate four editing prompts $p_{\text{edit}}^{1}$--$p_{\text{edit}}^{4}$ (Fig.~\ref{fig:prompt-edition}). The editing model takes as input the current RGB image together with a prompt, and produces an edited RGB output that predicts what happened a few seconds earlier. Each prompt progressively adds explicit constraints to guide the prediction: $p_{\text{edit}}^{1}$ preserves the original scene setup (lighting, angle, and colors) while introducing temporal consistency with the thermal trace; $p_{\text{edit}}^{2}$ enforces the presence of only one person in the scene; $p_{\text{edit}}^{3}$ removes the current position and renders only the person in the new past position/action; and $p_{\text{edit}}^{4}$ integrates both environment preservation and position replacement simultaneously. In all cases, the underlying action description is identical: \textit{``the person was sitting and holding the book''}. This progression isolates the effect of prompt structure on the editing model, showing that more structured prompts yield past-action predictions that are more temporally consistent, visually coherent with the scene, and faithful to the intended description.

\end{document}